\documentclass{article} 
\usepackage[preprint]{neurips_2024}

\usepackage{times}

\usepackage{amsmath,amsfonts,bm}

\def\eqref#1{equation~\ref{#1}}

\def\1{\bm{1}}

\DeclareMathAlphabet{\mathsfit}{\encodingdefault}{\sfdefault}{m}{sl}
\SetMathAlphabet{\mathsfit}{bold}{\encodingdefault}{\sfdefault}{bx}{n}

\usepackage{times}
\usepackage{latexsym}
\usepackage[T1]{fontenc}

\usepackage[utf8]{inputenc}

\usepackage{microtype}

\usepackage{inconsolata}
\usepackage{booktabs}
\usepackage{multirow}
\usepackage{longtable}

\usepackage{pifont}
\usepackage{tikz}
\usetikzlibrary{shapes.geometric, arrows}

\usepackage{amsmath} 
\usepackage{graphicx} 
\usepackage{color} 
\usepackage{xcolor} 
\usepackage{caption}
\usepackage{subcaption}

\definecolor{boxcolor}{RGB}{51,51,153}
\definecolor{lightgreen}{rgb}{0.56, 0.93, 0.56}
\usepackage{sectsty} 
\usepackage{amsmath}
\usepackage{amssymb}
\usepackage{mathtools}
\usepackage{amsthm}
\usepackage{float}
\usepackage{url}
\usepackage{array}
\usepackage{caption}
\usepackage{xcolor,colortbl}
\usepackage{listings}
\usepackage{soul}
\usepackage{wrapfig}
\usepackage{ragged2e} 

\usepackage{hyperref}  

\theoremstyle{plain}

\definecolor{carolinablue}{rgb}{0.6, 0.73, 0.89}

\title{ALLaVA: Harnessing  GPT4V-Synthesized Data for Lite Vision-Language Models} 

\author{Guiming Hardy Chen, Shunian Chen, Ruifei Zhang, Junying Chen, Xiangbo Wu, Zhiyi Zhang, \\ \textbf{Zhihong Chen}$^*$, \textbf{Jianquan Li}, \textbf{Xiang Wan}, \textbf{Benyou Wang}\thanks{Zhihong and Benyou are the corresponding authors}\\
Shenzhen Research Institute of Big Data\\
The Chinese University of Hong Kong, Shenzhen\\
\texttt{zhihongchen@link.cuhk.edu.cn}, \texttt{wangbenyou@cuhk.edu.cn} \\
\url{https://github.com/FreedomIntelligence/ALLaVA} \\
\url{https://huggingface.co/datasets/FreedomIntelligence/ALLaVA-4V} \\
}

\begin{document}
\newcommand{\basename}{ALLaVA} 
\newcommand{\shortname}[1]{ALLaVA-{#1}B} 

\newcommand{\needcite}{{\color{red}{[citation]}}}
\newcommand{\needupdate}{{\color{red}{[needs update]}}}

\newcommand{\superscript}[1]{$^{\text{#1}}$}

\maketitle
\begin{abstract}
\label{sec:abstract}
Large vision-language models (LVLMs) have shown premise in a broad range of vision-language tasks with their strong reasoning and generalization capabilities. However, they require considerable computational resources for training and deployment. This study aims to bridge the performance gap between traditional-scale LVLMs and resource-friendly lite versions by adopting high-quality training data. 
To this end, we propose a comprehensive pipeline for generating a synthetic dataset. The key idea is to leverage strong proprietary models to generate (i) fine-grained image annotations for vision-language alignment and (ii) complex reasoning visual question-answering pairs for visual instruction fine-tuning, yielding 1.3M samples in total. 
We train a series of lite VLMs on the synthetic dataset and experimental results demonstrate the effectiveness of the proposed scheme, where they achieve competitive performance on 17 benchmarks among 4B LVLMs, and even perform on par with 7B/13B-scale models on various benchmarks.
This work highlights the feasibility of adopting high-quality data in crafting more efficient LVLMs. 
We name our dataset \textit{ALLaVA}, and open-source it to research community for developing better resource-efficient LVLMs for wider usage.

\end{abstract}


\section{Introduction}
\label{sec:intro}
Recent months have seen a flourish development of Large Vision-Language Models (LVLMs). These models are able to process visual and textual inputs, resembling the way humans process information in real-world scenarios. An LVLMs typically consists of two key components, which are a vision encoder and a Large Language Model (LLM). The former enables the model to \textit{see} and the latter empowers the model to \textit{process} and \textit{speak}. Therefore, LVLMs can not only perform traditional tasks such as captioning~\citep{agrawal2019nocaps,young2014image} and image-text retrieval~\citep{lin2015microsoft,young2014image}, but are also able to follow instructions from human and perform complex VQA tasks~\citep{li2023seedbench,liu2023mmbench,ge2023mllmbench,yu2023mmvet,fu2023mme}, making them a milestone to Artificial General Intelligence (AGI).

While LVLMs demonstrate their superior ability, they often require vast resources for training and deployment. For example, IDEFICS~\citep{idefics} is trained on hundreds of millions of data; Qwen-VL~\citep{bai2023qwenvl} and CogVLM~\citep{wang2023cogvlm} are trained on more than 1 billion samples. The huge cost impedes the democratization of LVLMs.
To make LVLMs more portable, several works resort to develop lite LVLMs~\citep{chu2023mobilevlm,zhu2024llava}. 
Although these models are more friendly to users with scarce computation resources, they are accompanied by a loss of performance to some certain extent, manifested by the performance gap between normal-sized LVLMs and lite ones.

Previous works have shown the potential to improve the model performance through high-quality data~\citep{chen2023sharegpt4v,wang2023believe,li2023silkie}.
Therefore in this paper, we investigate a naturally arisen question: \textit{Can scaling up high-quality data fill the performance gap between normal-sized LVLMs and lite ones?}
To answer it, we curate high-quality images from LAION~\citep{schuhmann2021laion400m} and Vision-FLAN~\cite{visionFlan2023}. 
We then propose a data generation framework \textit{Captioning-then-QA}, which is applied to prompt GPT-4V to generate a large-scale vision language dataset, consisting of high-quality captions, instructions and answers. 
We perform ablated experiment on existing LVLMs to show the efficacy of our dataset, and train several models based on Phi2\footnote{\url{https://huggingface.co/microsoft/phi-2}}, StableLM-2-1.6B~\citep{StableLM-2-1.6B} and Phi-3-mini~\citep{abdin2024phi3}, which are superb 4B-scale LLMs.
Benefited from our high-quality training data, we observe significant advancement of cognition and perception ability on our models.
We name our dataset and resulted models as \textit{ALLaVA}, \textit{\textbf{A} {\tiny{\textbf{L}ite} } \textbf{La}nguage and \textbf{V}ision \textbf{A}ssistant}.
We hope that ALLaVA can assist open-source community in democratizing powerful commercial models such as GPT-4V, and developing better LVLMs for wider application.


Our contributions are as follow: (i) We open-source the largest synthetic dataset ALLaVA for LVLM training. The dataset consists of 1.3M diverse samples with fine-grained captions, complex instructions and detailed answers generated by GPT-4V, and high-resolution images curated from different sourcess; (ii) We release a series of 4B-scale LVLMs trained on ALLaVA. These models perform on par with larger models on multiple benchmarks, demonstrating the superiority of the proposed dataset.



\begin{figure*}[t]
    \centering
    \vspace{-20pt}
    \includegraphics[width=.7\textwidth,keepaspectratio]{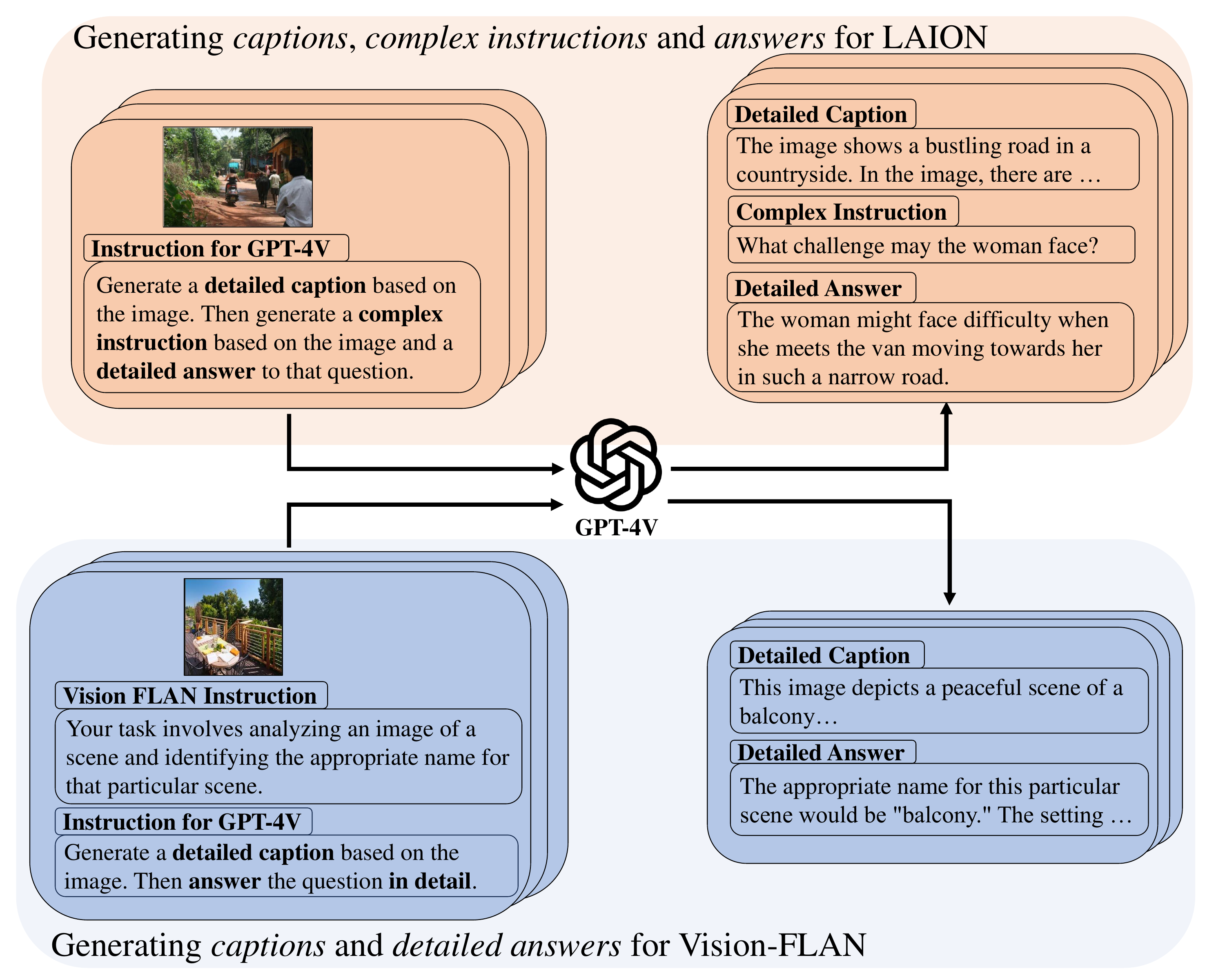}
    \caption{Pipeline for scaling up high-quality data. Prompts in the figure are shown for demonstration purpose. See the detailed prompt in Appendix~\ref{app:prompt_laion} and~\ref{app:prompt_vflan}.}
    \label{fig:pipeline}
\end{figure*}

\section{Rethinking Existing LVLMs}

Guided by the principle ``garbage in, garbage out'' which highlights the importance of input data quality, our approach reevaluates multimodal language models from a data-centric perspective. Within this framework~\footnote{We reassess the widely adopted solution (i.e. LLaVA~\cite{liu2023improved}) by scrutinizing both alignment and visual instruction tuning stages, with particular attention to potential issues.}, we focus on two primary strategies: \textit{alignment} and \textit{visual instruction tuning}.
The former is primarily dedicated to assisting language models in discerning visual objects and augmenting their visual reasoning capabilities. Meanwhile, the latter focuses on empowering LVLMs to generalize across a broader spectrum of instructions, particularly for those involving visual input. 




\subsection{On the Alignment}
\textbf{Image-text Alignment is rather Coarse-grained~~~}
Existing work tends to use caption data (i.e., images and its textual description)  to align images and texts in language models.
Popular large-scale caption datasets~\citep{schuhmann2022laion5b,schuhmann2021laion400m,sharma2018conceptual,changpinyo2021conceptual,ordonez2011im2text}
consist of short and course-grained captions, which introduces noisy signals and hinder the vision-language alignment process.
To improve their quality, BLIP~\citep{li2022blip} introduces CapFilt which is trained on human-annotated COCO~\citep{lin2015microsoft} dataset to generate higher-quality captions and remove unsatisfactory ones. 
LLaVA~\citep{liu2023visual} instead adopts the text-only GPT-4~\citep{openai2023gpt4} to directly generate  visual conversations using COCO annotations, but the detailedness of descriptions is bounded by human-annotation and is costly to scale up.
Besides, it is found that the cross-modal association in COCO image-text pairs is limited~\citep{parekh2020crisscrossed}, questioning the effectiveness of curating high-quality data on top of COCO. Therefore, we need a more reasonable and scalable approach for obtaining high-quality caption data.



\subsection{On Visual Instructions}
\label{sec:instruction_issues}
\textbf{Questions are Relatively Simple~~~}
Taking Vision-FLAN~\citep{visionFlan2023} as example, which comprises 191 VQA tasks across 101 datasets, its questions are relatively simple compared to WizardLM~\cite{xu2023wizardlm}.
As stated by WizardLM~\cite{xu2023wizardlm}, complex questions (or called `\textit{instruction}') are beneficial for language models, especially in terms of instruction following. Morever, current visual instruction tuning datasets focus more on improving fundamental abilities than on more advanced ones such complex reasoning.
For example, Visual Genome~\citep{krishna2016visual} consists of bounding-box locating questions, OCRVQA~\citep{mishra2019ocr} contains simple text recognition task for book covers, and TextVQA~\citep{singh2019towards} asks to generate one-sentence caption for each image.

\textbf{Answers are Short and Uninformative~~~}
Moreover, although the answers in  Vision-FLAN are manually annotated by human, they often consist of short word or phrases without format prompts. Some answers are even incomplete as a sentence (\textit{e.g.}, without a period or capitalizing the first letter). Directly learning such outputs would hinder the model performance~\citep{liu2023improved}, manifesting the need of polishing or regenerating the answers to Vision-FLAN instructions.

\section{Methodology of ALLaVA}
\label{sec:method}


\subsection{Motivation for Lite LVLMs}
Lite LVLMs are gaining increasing popularity since they cost less to train and deploy.
For training, a 7B LLaVA-architecture model takes less than 14 hours to be trained on 1M data with 8*A100 40G GPUs~\citep{liu2023improved}. The training time decreases linearly with number of trainable parameters, which means it takes only less than 7 hours to train a Phi2-2.7B backbone under the same settings.
The deployment cost of lite LVLMs is much smaller than normal-size ones. Using quantization techniques, one can fit a 2.7B LVLM into a 8GB-RAM mobile phone and conduct inference~\citep{chu2023mobilevlm} with decent performance, indicating a promising future of lite LVLMs.

\subsection{The Philosophy of ALLaVA}

\textbf{Harnessing High-Quality Data for Scale Compensation}
While lightweight LVLMs offer  advantages over their normal-size counterparts in terms of computational cost, they might experience performance drops due to their reduced number of parameters. To enhance the effectiveness of lite LVLMs while preserving their efficiency, our goal is to construct high-quality datasets that can compensate for the diminished capacity inherent in lite LVLMs when compared to their normal-size counterparts.

\textbf{Data Synthesis using a \textbf{Captioning-then-QA} Fashion}
To generate high-quality captions and VQAs, we propose to distill a caption and a QA pair for an image within a single session, see Figure~\ref{fig:overall_prompt}. Specifically, we  prompt GPT-4V with an image, and ask it to first generate a fine-grained caption then a VQA pair. By doing so, the whole data synthesis  procedure including three stages: \textit{captioning}, questioning and  answering, which are described in Section~\ref{sec:pipeline}.

\begin{figure}[h]
\lstset{
 backgroundcolor=\color[RGB]{245,245,244},
 breaklines=true,
 basicstyle=\ttfamily\small
}
\begin{lstlisting}
### You are an excellent image describer and questioner
### You have three tasks in total
#### Your first task is to describe the given image as detailed as possible
#### Your second task is to ask a complex question ...
#### Your third task is to answer the question you raised solely based on the given image
\end{lstlisting}
    \caption{An example prompt snippet for the overall pipeline in a captioning-then-
    QA fashion. It includes (1) \textit{captioning}, (2) \textit{questioning}, and (3) \textit{answering}.}
    \label{fig:overall_prompt}
\end{figure}

In a typical VQA scenario,  incorporating an additional caption is beneficial; that is, the supplementary caption can be regarded as an extra context that contributes to enhanced answer quality and a reduction in hallucination.
Since \textit{image embeddings} and \textit{caption} serve as \textit{implicit} and \textit{explicit} expression of images, respectively, the generation of answers can be based on the two types of expressions instead of just the former. 
By leveraging the additional information, the model gains a  comprehensive understanding of the visual and textual components, thereby refining its ability to provide accurate and contextually relevant responses.
Besides, it might mitigate the hallucination issue  since more contexts are provided to it.

\subsection{Data Curation Pipeline}
\label{sec:pipeline}

In this subsection, we introduce the four stages: \textit{image selection}, \textit{captioning}, \textit{questioning} and  \textit{answering} in Section~\ref{sec:image_sources}, \ref{sec:stage1}, \ref{sec:stage2} and \ref{sec:stage3}, respectively.

\subsubsection{Stage 0: Image Selection}
\label{sec:image_sources}
We select two sources for images in data  synthesis: LAION~\citep{schuhmann2021laion400m} and Vision-FLAN~\citep{visionFlan2023} (VFLAN in short in the rest of this paper).
\textbf{LAION} is a popular dataset for vision-language alignment since it contains diverse images that are crawled from webpages. Therefore, the image sources are also aligned with the real-world usages of end users.
To ensure image quality, we perform quality control and deduplication on these images, the process of which is detailed in Appendix~\ref{app:dedup_laion}.
\textbf{Vision-FLAN} is a dataset that integrates 191 VQA tasks across 101 open-source datasets. It comprises instructions that are vital for improving the foundation ability of LVLMs and can enhance the performance on vision-language benchmarks.

\subsubsection{Stage 1: Captioning}
\label{sec:stage1}
\begin{figure}[h]
\lstset{
 backgroundcolor=\color[RGB]{245,245,244},
 breaklines=true,
 basicstyle=\ttfamily\small
}
\begin{lstlisting}
You are an excellent image describer.

Your task is to first describe an image and then answer a question.

Your description should include details about the main subjects, background elements, colors, and any notable features. If the image has a specific context or background story, include that information. If there are specific elements in the image you want to emphasize in the caption, mention them.
...
\end{lstlisting}
    \caption{An example prompt snippet for captioning.}
    \label{fig:prompt_captioning}
\end{figure}

\textbf{Towards Fine-grained Captions~~~}
Figure~\ref{fig:prompt_captioning} shows a prompt snippet for captioning. It asks GPT-4V to keep an eye on multiple aspects of an image and describe the image as detail as possible.  
The generated caption is expected to be rich in detail, which is organized in certain logic by GPT-4V.

\subsubsection{Stage 2: Questioning}
\label{sec:stage2}

Since Vision-FLAN already contains diverse instructions, we retain its original instructions and do not perform question generation on it. We only generate questions for images sourced from LAION.

\begin{figure}[h]
\lstset{
 backgroundcolor=\color[RGB]{245,245,244},
 breaklines=true,
 basicstyle=\ttfamily\footnotesize
}
\begin{lstlisting}
...
#### Your second task is to ask a complex question that requires close inspection of the image and strong reasoning ability to answer, you should ask FIVE candidate questions in different aspects and diverse ways, then RANDOMLY choose one of them to answer.
...
\end{lstlisting}
    \caption{An example prompt snippet for question generation.}
    \label{fig:prompt_question}
\end{figure}



\textbf{Prompting for Complex and Diverse Questions~~}
As argued from Sec.~\ref{sec:instruction_issues}, we aim to generate complex questions that probably involves complex reasoning. This is inspired by WizardLM~\citep{xu2023wizardlm}, but we implement complex instructions using a light-weight prompting instead of original instruction evolution. 
We demonstrate an example prompt for question generation in Figure~\ref{fig:prompt_question}. Besides requiring the \textit{complexity} of questions generated, we also expect \textit{diversity} of questions. Hence, we ask GPT-4V to generate five questions at a time and choose one to answer.

\subsubsection{Stage 3: Answering}
\label{sec:stage3}

\begin{figure}[h]
\lstset{
 backgroundcolor=\color[RGB]{245,245,244},
 breaklines=true,
 basicstyle=\ttfamily\footnotesize
}
\begin{lstlisting}
...
Your answer should provide relevant information to the question and demonstrate the process of solving the question.
...
\end{lstlisting}
    \caption{An example prompt snippet for question answering.}
    \label{fig:prompt_answer}
\end{figure}

\textbf{Detailed Answers~~~}
Figure~\ref{fig:prompt_answer} shows an example prompt for generating answer to a given question. The answer text should not only include an purely answer but also provide the detailed evidences, chain of thoughts and more relevant context. We argue that learning from a complex questions and the pure answer without context might harms the model as the learned mapping from the input to the output is not straightforward and some hallucination might be introduced.


For LAION, we strictly followed the aforementioned data generation pipeline. For VFLAN, considering the diversity of its original questions, we decided to retain its questions but regenerate the answers. As a result, the questions in the LAION dataset compensated for the simplicity of the questions in the VFLAN dataset, while the original questions in VFLAN supplemented the overall diversity of the dataset's questions.


\subsection{On the Ethics}
\label{sec:ethics}
Ethical considerations are imperative.
For example, it is crucial to address prompts that involve traditionally biased elements such as gender and races when describing specific occupations. 
In our prompt, we require that any question attempting to elicit responses involving the disclosure of personal information or encourages discriminatory judgments should be promptly refused (see Appendix~\ref{app:prompt_laion}). 
Upholding ethical standards is essential in maintaining a responsible and inclusive approach to language generation, fostering an unbiased environment in the information provided.


% ...

% ...



\subsection{On the Effectiveness of Caption-then-QA Pipeline}
\label{sec:vflan_manual_inspection}


To demonstrate the efficacy of our data generation strategy, we uniformly choose 100 samples from different categories of VFLAN and directly generate the answer for each question without captioning first. Then we perform a manual inspection of the accuracy of the answers. Table~\ref{tab:vflan_manual_inspection} summarizes the results for manual inspection. 

\begin{wraptable}{r}{6cm}
\vspace{-15pt}
\centering
\caption{Results of manual inspection on 100 samples from VFLAN.}
\resizebox{\linewidth}{!}{

\begin{tabular}{l|c}
\toprule
   Answers from  & Accuracy (\%) \\ \midrule
   {\color{gray}\textit{VFLAN gt}} & {\color{gray}88.0} \\
   \textit{Direct Answer}  & 78.0 \\
   \textit{\textbf{Caption-then-Answer (ours)}} & 84.0 \\
\bottomrule
\end{tabular}
}
\label{tab:vflan_manual_inspection}
\end{wraptable}
Our \textit{Caption-then-Answer} strategy exceeds \textit{Direct Answer} by 6\% in accuracy, though falling behind \textit{VFLAN gt} by 4\%. 
We demonstrate the superiority of our strategy in Appendix~\ref{app:vflan_manual_inspection}, where an example is shown to compare \textit{Caption-then-Answer} and \textit{Direct Answer}. 
The manual verification indicates the efficacy of our Caption-then-Answer curation pipeline.


\subsection{The Resulted Dataset}



\begin{wraptable}{r}{6cm}
\centering
\caption{Summary of ALLaVA dataset.}
\resizebox{\linewidth}{!}{
    \begin{tabular}{lll}
    \toprule
    Subsets & \#Ex. & Total  \\ \midrule
    \rowcolor{lightgray}
    \textit{\textbf{{ALLaVA-Caption-4V}}} && \\
    \textbf{{ALLaVA-Caption-LAION-4V}}  & 469K & \multirow{2}{*}{664K} \\
    \textbf{{ALLaVA-Caption-VFLAN-4V}}  & 195K &  \\
    \midrule
    \rowcolor{lightgray}
    \textit{\textbf{ALLaVA-Instruct-4V}} & & \\
    \textbf{ALLaVA-Instruct-LAION-4V} & 469K & \multirow{2}{*}{663K} \\
    \textbf{{ALLaVA-Instruct-VFLAN-4V}} &  194K &  \\ 
    \bottomrule
    \end{tabular}
}
\label{tab:distilled_datasets}
\end{wraptable}
As seen in Table~\ref{tab:distilled_datasets}, we build two large-scale synthetic datasets following our data generation pipeline: image caption and visual instruction data.

 \textbf{LAION~~~}
The upper part of Figure~\ref{fig:pipeline} illustrates the pipeline for distilling a fine-grained caption and a complex VQA for the same image within one prompt using GPT-4V~\citep{gpt4v}, which is so far the most powerful LVLM developed by OpenAI.
We use a subset of 469K images from LAION~\citep{schuhmann2021laion400m}, which contains diverse images crawled from webpages. We name the distilled caption set as \textbf{{ALLaVA-Caption-LAION-4V}}, and the VQA set as \textbf{{ALLaVA-Instruct-LAION-4V}}. 
See the detailed prompt in Appendix~\ref{app:prompt_laion} and samples in Appendix~\ref{sec:data_example}.

\textbf{VFLAN~~~}
The lower part of Figure~\ref{fig:pipeline} showcases the pipeline for distilling a fine-grained caption and a detailed answer to a given instruction for the same image within one prompt using GPT-4V. We name the distilled caption set as \textbf{{ALLaVA-Caption-VFLAN-4V}}, and the VQA set as \textbf{{ALLaVA-Instruct-VFLAN-4V}}. See the detailed prompt in  Appendix~\ref{app:prompt_vflan} and samples in Appendix~\ref{sec:data_example}.

\begin{figure}[t]
    \centering
    \vspace{-20pt}
    \includegraphics[width=1\textwidth,keepaspectratio]{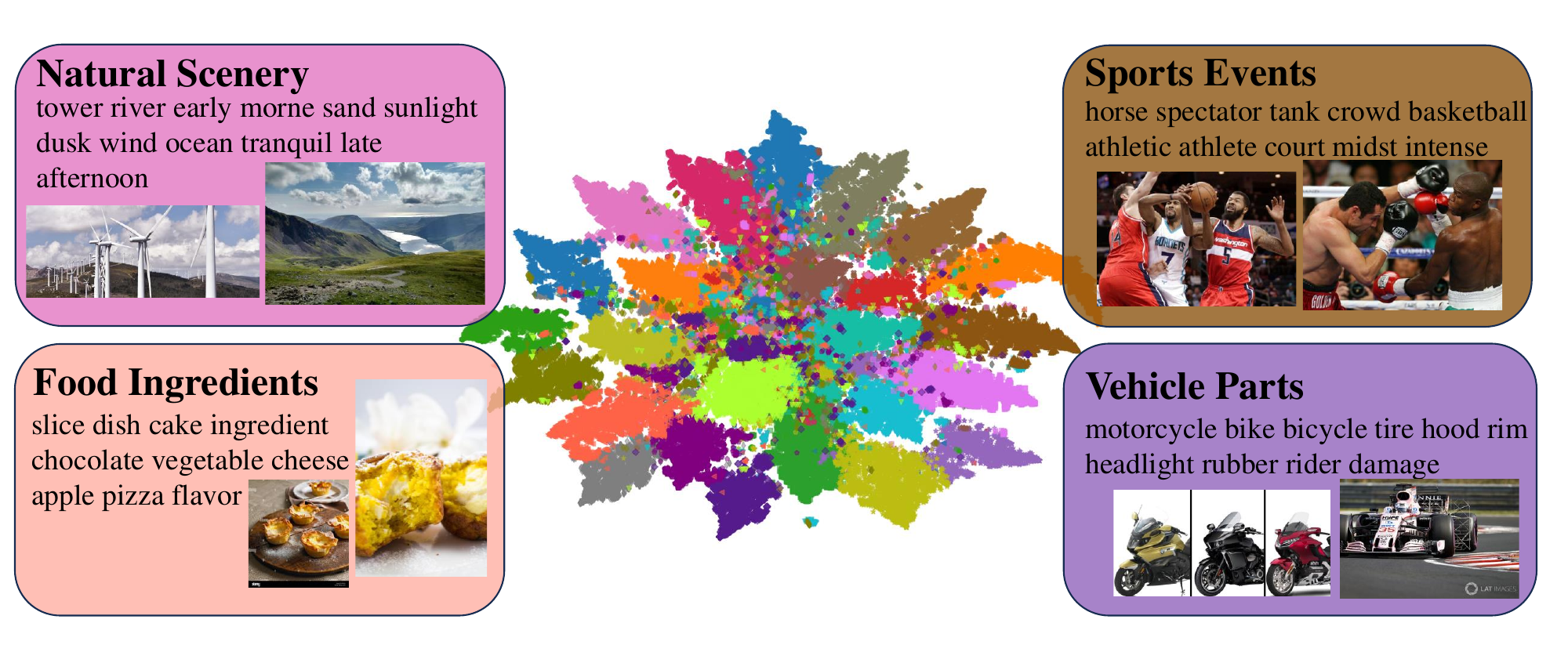}
    \caption{Visualization of LDA results using t-SNE. Four clusters are chosen for a detailed visualization with topic name, top-10 key words and sample images.}
    \label{fig:topic_analysis}
\end{figure}

\section{Exploring ALLaVA}
\label{sec:explore_allava}

\begin{figure}[t]
\begin{minipage}{0.45\textwidth}
\centering
\vspace{-20pt}
\captionof{table}{Comparison of ALLaVA-Caption, ALLaVA-Instruct and their respective existing counterpart datasets. {We only consider instruction data in LVIS-4V.}}
\resizebox{\linewidth}{!}{
\begin{tabular}{lcc}
\toprule
\rowcolor{lightgray}
& ALLaVA-Caption & ShareGPT4V \\
\#Samples & 664K & 102K \\
Resolution(w/h) & 891/770 & 534/451\\
Image Sources & Diverse & Diverse \\
\midrule
\rowcolor{lightgray}
& ALLaVA-Instruct & LVIS-4V \\
\#Samples & 663K & 110K \\
Resolution(w/h) & 891/770 & 576/484 \\
Image Sources & Diverse & COCO~\citep{lin2015microsoft} \\
\bottomrule
\end{tabular}
}
\label{tab:basic_stats}
\end{minipage}%
\hfill{}
\begin{minipage}{0.52\textwidth}
\centering
\includegraphics[width=\linewidth,keepaspectratio]{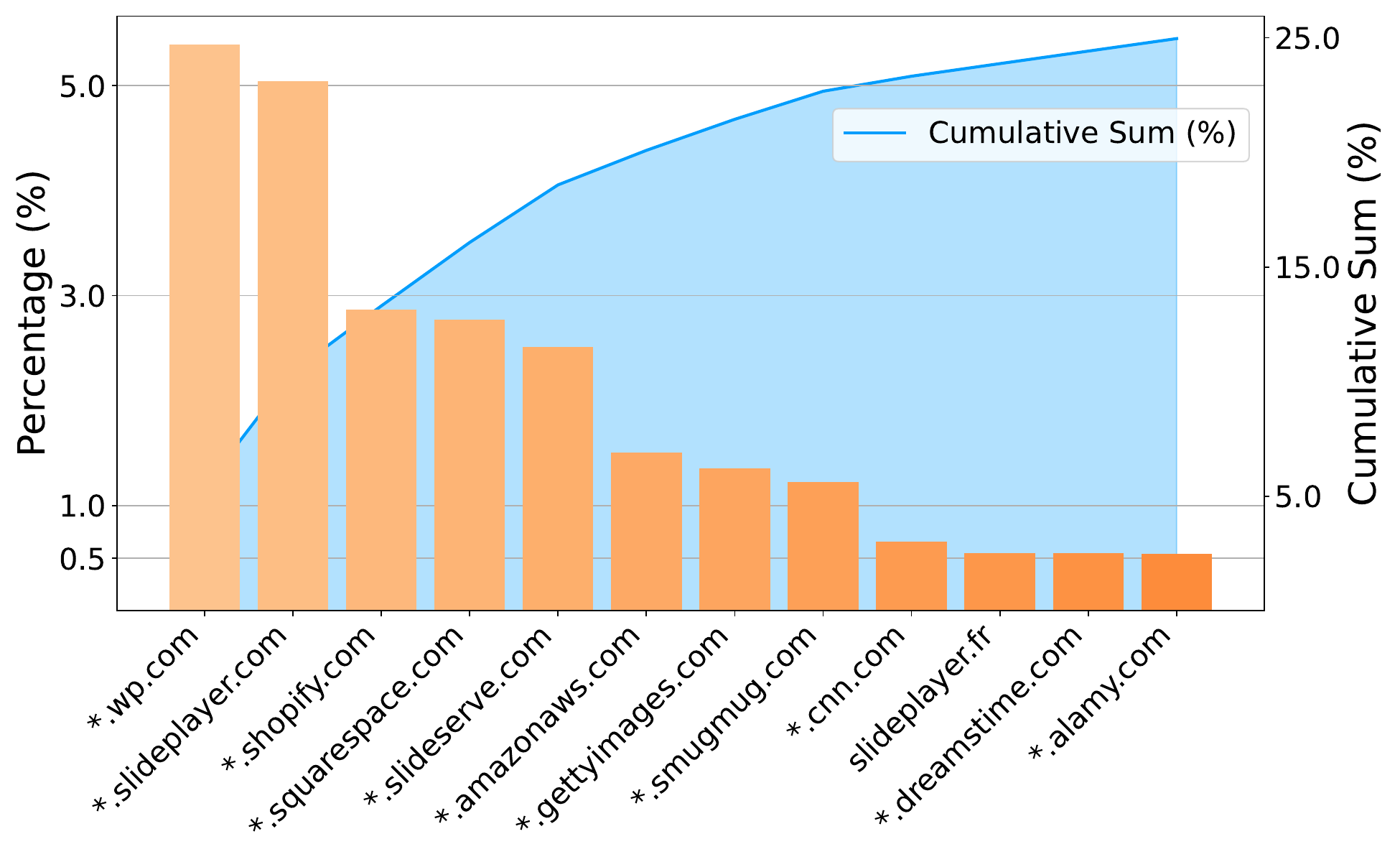}
\caption{Distribution of top-12 URLs in ALLaVA-LAION. 
The shaded blue area shows the cumulative sum of percentages.}
\label{fig:url_distribution}
\end{minipage}

\end{figure}

\textbf{Basic Statistics~~~}
As depicted in Table~\ref{tab:basic_stats}, the ALLaVA dataset comprises 664K samples, with average image resolutions of 891 by 770 pixels, sourced from a diverse range of images. 
The ALLaVA dataset is divided into two subsets: ALLaVA-Caption and ALLaVA-Instruct, which share the same set of images. 
It is important to note that ALLaVA surpasses other public datasets in multiple dimensions, such as \#samples, average resolutions, etc.

\textbf{Diverse Sources of Images in ALLaVA-LAION~~~}
To show the diverse source of images in ALLaVA-LAION, we analyze the top-level domains of image URLs. 
We find that the 469K images are from 122K unique domains. We also illustrate the distribution of top-12 domains in Figure~\ref{fig:url_distribution}.
The traced URLs span encompass multiple types of websites, including news (\textit{e.g., }CNN), high-res templates (\textit{e.g., }WordPress, Squarespace), merchandise (\textit{e.g., }Amazon AWS, Shopify), high-res photographs (\textit{e.g., }SmugMug, alamy), etc. 
These commercial websites need to deliver decent visual experience for viewers, guaranteeing superior image quality. Meanwhile, the images on these websites encapsulate a variety of daily life scenarios, thus increasing their naturalness and diversity.





\textbf{Topic Analysis}
To show the diversity of topics covered in the dataset, we run Latent Dirichlet Allocation (LDA)~\citep{blei2003latent} implemented by Mallet~\citep{McCallumMALLET} on ALLaVA-Caption-LAION-4V and ALLaVA-Caption-VFLAN-4V. 
Specifically we randomly draw a total of 100K samples from two sets, set the number of topics to 25 and train an Mallet-LDA model for 1K steps. Then for each of the 25 clusters, we feed the most relevant 10 words to GPT-4 (prompts in Appendix~\ref{app:prompt_gpt4_topic_summary}) and generate a summary of these words as the topic name of the cluster.
Figure~\ref{fig:topic_analysis} visualizes the 25 topic clusters with t-SNE~\citep{van2008visualizing}, where each color denotes a cluster. 
A full list of the 25 topics are shown in Appendix~\ref{app:list_of_topics}.
This experiment demonstrates the widespread of topics in ALLaVA, encompassing
Natural Scenery,
Fashion Accessories, 
Food Ingredients, 
Home Decor, 
Sports Events,
Vehicle Parts,
\textit{etc.}, which potentially improves the generalizability of our model.


\section{Experiments}
\label{sec:experiment}
\subsection{Effects of the ALLaVA Dataset on Existing LVLMs}

\begin{figure}[htpb]
    \centering
    \vspace{-30pt}
    \begin{subfigure}{.48\textwidth}
        \centering
        \includegraphics[width=\linewidth]{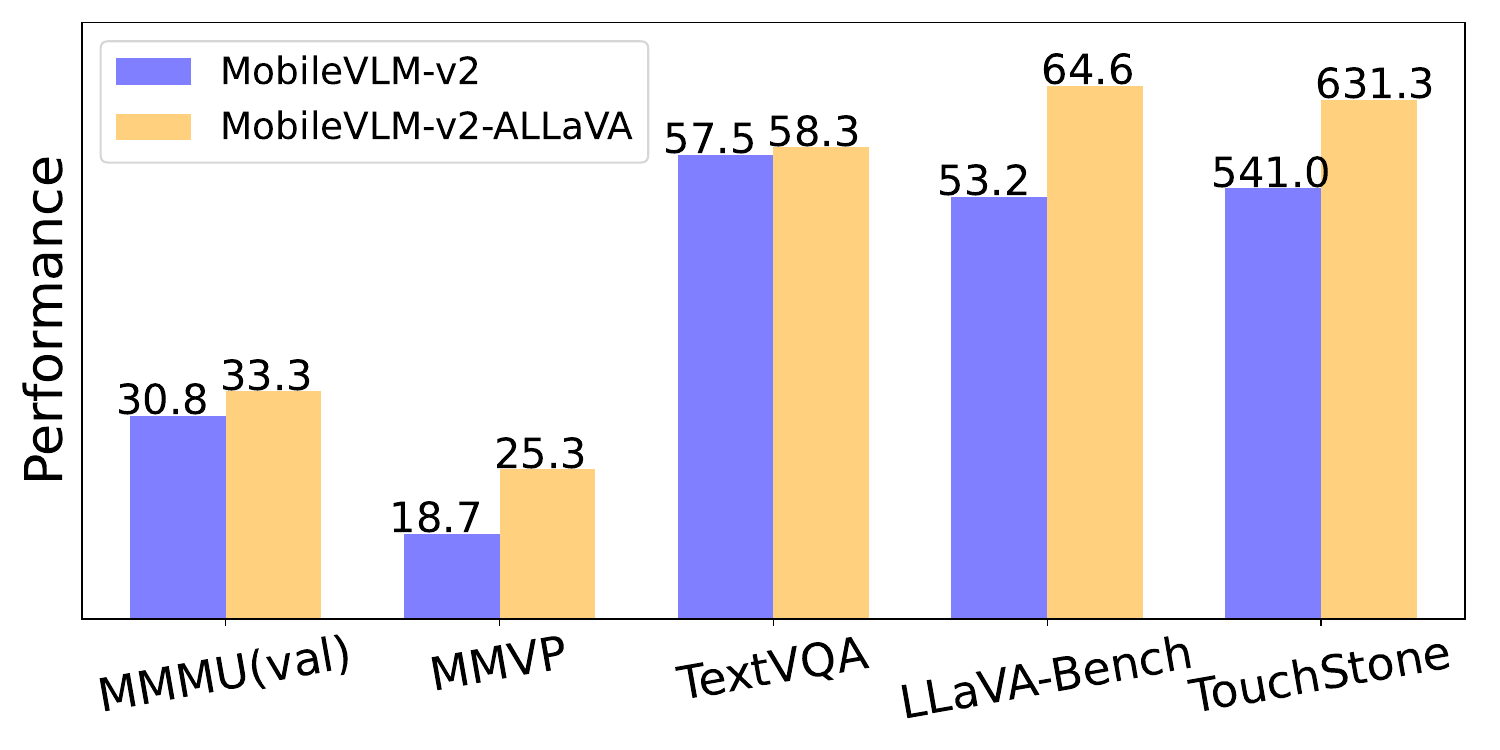} 
        \caption{MobileVLM-v2}
        \label{fig:ablation_lvlms_mobilevlm}
    \end{subfigure}%
    \hspace{1pt} 
    \begin{subfigure}{.48\textwidth}
        \centering
        \includegraphics[width=\linewidth]{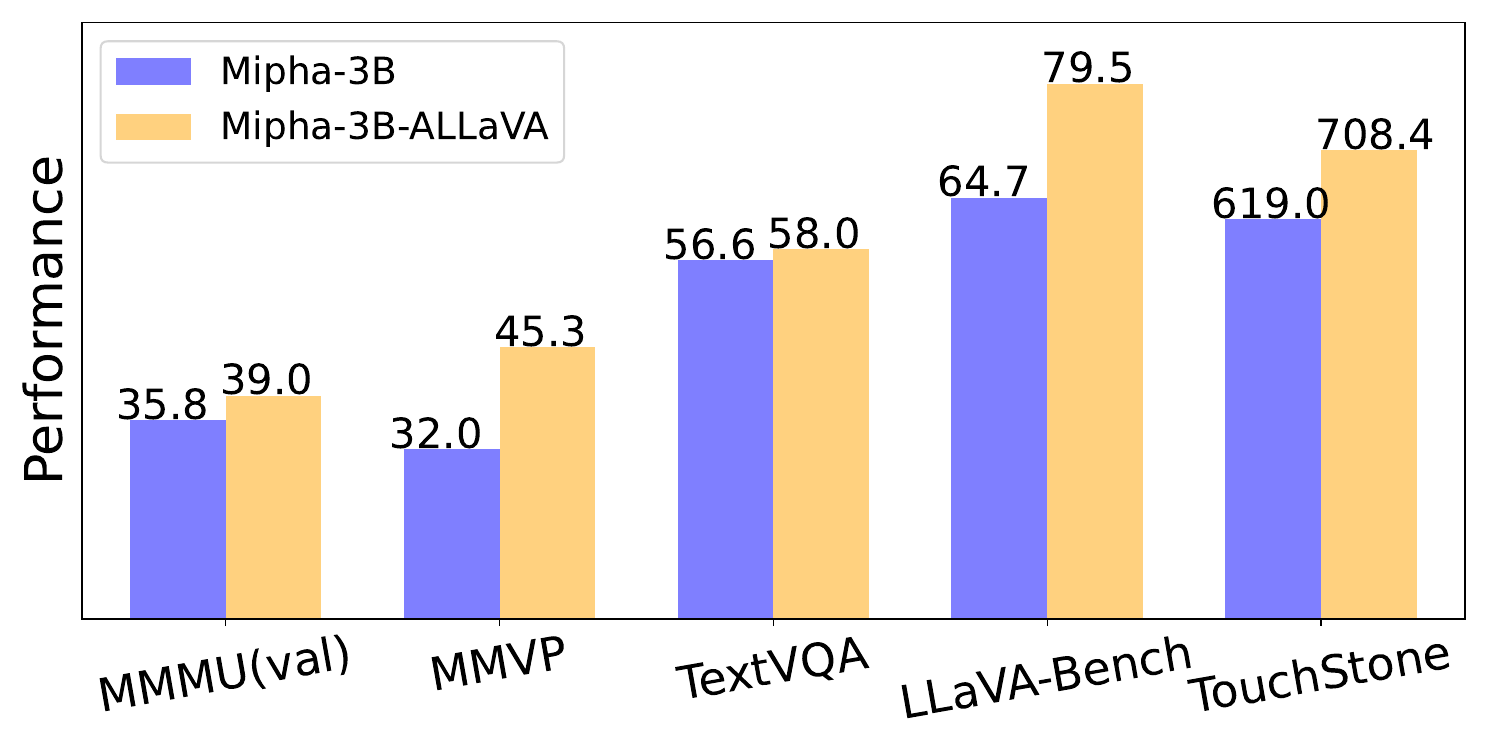} 
        \caption{Mipha-3B}
        \label{fig:ablation_lvlms_mipha}
    \end{subfigure}
    \caption{ Ablation of adopting ALLaVA in training MobileVLM-v2~\citep{chu2023mobilevlm} and Mipha-3B~\citep{zhu2024llava}. We use their respective source code for training our own versions of models.}
    \label{fig:ablation_lvlms}
\end{figure}

\subsubsection{Settings}
To show the effectiveness of ALLaVA, we perform ablated experiments on two popular lite VLMs, MobileVLM-v2~\citep{chu2023mobilevlm} and Mipha-3B~\citep{zhu2024llava}. 
MobileVLM-v2 is based on MobileLLaMA-2.7B-Chat~\citep{liu2024mobilellm} with CLIP~\citep{radford2021learning} as the vision encoder. Mipha-3B is based on Phi-2 and SigLIP~\citep{zhai2023sigmoid} is used to encode images. We add 
Both models adopt a two-stage training paradigm, which our dataset fits in. In the ablated setting, we solely add ALLaVA data to each stage and measure model performances on MMMU~\citep{yue2023mmmu}, MMVP~\citep{tong2024eyes}, TextVQA~\citep{singh2019towards}, LLaVA-Bench (In-the-Wild)~\citep{liu2023visual} and TouchStone~\citep{bai2023touchstone}. 

\subsubsection{Results and Analysis}
Figure~\ref{fig:ablation_lvlms_mobilevlm} and~\ref{fig:ablation_lvlms_mipha} show the results for MobileVLM-v2 and Mipha-3B, respectively.
With the inclusion of ALLaVA data in both stages, performances of both models on all five benchmarks are boosted in various degree. Notably, model performances are largely advanced on MMVP, LLaVA-Bench and TouchStone. 
MMVP and TouchStone focus on testing basic perception ability of models, whereas LLaVA-Bench is designed to evaluate their cognition and reasoning ability in responding to complex queries. A significant boost on these datasets demonstrate the efficacy of incorporating ALLaVA to comprehensively enhance the performance of lite VLMs.





\subsection{Comparison with Other LVLMs}
\subsubsection{Settings}

\textbf{Architecture~~~}
Having proved the efficacy of ALLaVA on existing LVLMs, we intend to develop a series of LVLMs with different backbones to show the potential of ALLaVA.
Our model architecture is identical to LLaVA-v1.5~\citep{liu2023improved}, which consists of a vision encoder (CLIP-ViT-L/14@336), a projector and an LLM (Phi-2-2.7B, StableLM-2-1.6B~\citep{StableLM-2-1.6B} and Phi-3-mini-128K~\citep{abdin2024phi3}). 
We adopt two-stage training following LLaVA-v1.5~\citep{liu2023improved}. We train the projector and LM backbone, and freeze the vision encoder at both stages.
More detailed training hyperparamters are shown in Appendix~\ref{app:training_details}.

\textbf{Training Data~~~}
We detail the training data in Appendix~\ref{app:training_data}, consisting of 149K text data, 795K caption data and 1,372K VQA data.
The textual dataset~\citep{chen2023phoenix} consists of instructions from~\citep{xu2023wizardlm} and answers generated by GPT4-Turbo (originally GPT3.5-Turbo). 
At pretraining stage, we randomly mix up one copy of caption data and \textit{two} copies of textual data. The two copies of textual data aids to equip a base LLM with instruction following ability.
At finetuning stage, we randomly mix up one copy of VQA data and one copy of textual data. The textual data are added to mitigate the catastrophic forgetting issue of LLM during visual instruction finetuning~\citep{bai2023qwenvl}.

\textbf{Evaluation~~~}
We evaluate our model performance on 1 textual benchmark and 14 multimodal benchmarks. We divide multimodal benchmarks into three subsets, \textit{Perception}, \textit{Cognition} and \textit{Comprehensive} to differentiate the tested abilities. A full list of benchmarks, a brief introduction and metrics are shown in Appendix~\ref{sec:benchmarks}.

\subsubsection{Results and Analysis}
Table~\ref{tab:quantitative_results} shows a comparison between ALLaVA models and other similar-scale and large-scale VLMs. 
In general, ALLaVA models outperform 4B-scale VLMs on most of the 17 benchmarks. Our models also perform on par with 7B or larger-scale VLMs

\textbf{Multimodal Perception~~~}
The perception ability is largely dependent on 1) vision encoder with fine-grained representations and 2) training datasets with fine-grained annotations. ALLaVA models adopts CLIP as vision encoder, which yields features with more course granularity compared to SigLIP's. This shortcoming is compensated with ALLaVA data. Thus, ALLaVA models are still able to outperform other 4B-scale models on several Perception benchmarks.

\textbf{Multimodal Cognition~~~}
The performance of ALLaVA models outperform other 4B-scale models on 6/7 cognition benchmarks. This result demonstrate that ALLaVA dataset is able to boost the cognitive ability, ore specifically on Math, complex reasoning and multimodal multi-discipline knowledge.

\textbf{Language Ability~~~}
We also include a textual benchmark to test the language ability of LVLMs after visual instruction tuning. 
All ALLaVA models score significantly higher than 4B-scale models. ALLaVA-Phi2, Mipha-3B and TinyLLaVA share the same backbone, but their performance gap is distinctive (49.4 v.s. 16.2 and 15.6). ALLaVA-Phi3 performs the best among the three lite ALLaVA models. The results indicate the importance of choosing a strong language backbone, as well as adding textual data to avoid catastrophic forgetting during instruction tuning~\citep{bai2023qwenvl}.

\begin{table*}[t]
\centering
\vspace{-20pt}
\footnotesize
\caption{
Evaluation results on 1 textual benchmark and 16 multimodal benchmarks. * denotes the results of our evaluation.
Vicuna-80~\citep{vicuna2023}; 
GQA~\citep{hudson2019gqa};
HallB: HallusionBench~\citep{guan2023hallusionbench};
MME (Perception and Cognition)~\citep{fu2023mme};
MMVP~\citep{tong2024eyes};
TS: TouchStone~\citep{bai2023touchstone}; 
VQA$^T$: TextVQA~\citep{singh2019towards};
MV: MathVista~\citep{lu2024mathvista};
MM-Vet~\citep{yu2023mmvet};
MMMU$^{val}$: validation set of MMMU~\citep{yue2023mmmu};
SQA$^{I}$: ScienceQA-Image~\citep{lu2022learn}; 
LLaVA$^{W}$: LLaVA-Bench (In-the-Wild)~\citep{liu2023visual}; 
MB: MLLM-Bench~\citep{ge2023mllmbench};
MMB: MMBench~\citep{liu2023mmbench};
SEED\(_{img}^{v1}\): image set of SEED-Bench-v1~\citep{li2023seedbench}; 
\textbf{Bold numbers} are the best results among all 4B-scale LVLMs.
}
\resizebox{\textwidth}{!}{
\setlength{\tabcolsep}{2pt}
\begin{tabular}{ l   c c c c c c c c c c c c c c c c c}
\toprule
\vspace{2pt} 
\multirow{1}{*}{Model} & \multicolumn{17}{c}{Benchmarks}  \\
\cmidrule{2-18}
& \multicolumn{1}{c}{Text} & \multicolumn{6}{|c|}{Perception} & \multicolumn{7}{|c|}{Cognition} & \multicolumn{3}{|c|}{Comprehensive} \\
\cmidrule{2-18}
& Vicuna-80 & GQA & HallB & MME-P & MMVP & TS  &  VQA$^T$ & MME-C & MV & MM-Vet & MMMU\(^{val}\) &  SQA$^I$ & LLaVA\(^{W}\) & MB & MMB-en & MMB-cn & SEED\(_{img}^{v1}\) \\
\midrule
\rowcolor{lightgray}
\multicolumn{18}{c}{\textit{\textbf{Large VLMs}}}   \\
BLIP-2~\citep{li2023blip2} & - & - & - & - & - & - & - & - & - & 22.4 & 34.4 & - & - & 3.0* & - & - & 49.7\\
InstructBLIP~\citep{dai2023instructblip} & - & 49.5 & - & - & - & 552.4 & 50.7 & - & - & 25.6 & - & - & 58.2 & - & 44.0 & - & -\\
Qwen-VL-Chat~\citep{bai2023qwenvl} & - & 57.5 & - & 1487.6 & - & - & 61.5 & 360.7 & - & 31.1 & - & 68.2 & - & - & 60.6 & 56.7 & 65.4\\
LLaVA-1.5-7B~\citep{liu2023improved} & 13.8* & 62.0 & 36.6* & 1504.4* & 24.7* & 594.9* & 58.2 & 324.6* & 25.0* & 31.1 & 35.1* & 66.8 & 65.4 & 23.0*& 64.3 & 58.3 & 66.1\\
LLaVA-1.5-13B~\citep{liu2023improved} & 22.5 & 63.3 & 36.5* & 1531.3 & 38.0* & 617.7* & 61.3 & 295.4 & 28.3* & 35.4 & 34.4* & 71.6 & 72.5 & - & 67.7 & 63.6 & 68.2\\
LVIS-7B~\citep{wang2023believe} & - & 62.6 & - & - & - & - & 58.7 & - & - & 31.5 & - & - & 67.0 & 29.0* & 66.2 & - & -\\
LVIS-13B~\citep{wang2023believe} & - & 63.6* & - & - & - & - & 62.5* & - & - & 37.4* & - & - & 71.3* & - & 68.0* & - & -\\
ShareGPT4V-7B~\citep{chen2023sharegpt4v} & 13.8* & 63.3 & 36.0* & 1540.1* & 34.0* & 637.2* & 60.4 & 346.1* & 24.7* & 37.6 & 35.4* & 68.4* & 72.6 & 30.2* & 68.8 & 61.0* & 69.7\\
ShareGPT4V-13B~\citep{chen2023sharegpt4v} & 17.5* & 64.8 & 39.0* & 1576.1* & 35.3* & 648.7* & 62.2 & 309.3* & 28.8* & 43.1 & 35.6* & 70.0* & 79.9 & 35.5* & 71.2 & 61.7* & 70.8\\
\midrule

\rowcolor{lightgray}
\multicolumn{18}{c}{\textit{\textbf{4B-scale Lite VLMs}}}   \\
MobileVLM-v2~\citep{chu2023mobilevlm} & 5.0* & 61.1 & 30.8* & 1440.5 & 18.7* & 541.0* & 57.5 & 261.8* & 28.3* & 26.1* & 30.8* & 70.0 & 53.2* & 15.7* & 63.2 & 43.2* & 64.5*\\
Mipha-3B~\citep{zhu2024llava} & 16.2* & \textbf{63.9} & 34.3* & \textbf{1488.9} & 32.0* & 619.0* & 56.6 & 285.0* & 27.8* & 33.5* & 35.8* & 70.9 & 64.7* & 23.1* & \textbf{69.7} & 42.9* & \textbf{71.2*} \\
TinyLLaVA~\citep{jia2024tinyllava} & 15.6* & 62.1 & 37.2* & 1465.5* & 33.3* & 663.5* & \textbf{60.3} & 281.1* & 30.3* & 37.5 & 38.4 & \textbf{73.0} & 70.8* & 29.8* & \textbf{69.7*} & 42.8* & 70.4*\\
\midrule

\rowcolor{lightgray}
\multicolumn{18}{c}{\textit{\textbf{Ours}}}   \\\rowcolor{carolinablue!60}

\textbf{ALLaVA-Phi2} & 49.4 & 48.8 & 24.8 & 1316.2 & \textbf{36.0} & 632.0 & 49.5 & 301.8 & 27.4 & 32.2 & 35.3 & 67.6 & 69.4 & 43.6 & 64.0 & 40.8 & 65.2\\\rowcolor{carolinablue!60}
\textbf{ALLaVA-StableLM2} & 38.8 & 49.8 & 25.3 & 1311.7 & 34.0 & 655.2 & 51.7 & 257.9 & 27.7 & 31.7 & 33.3 & 64.7 & \textbf{72.0} & 39.3 & 64.6 & 49.8 & 65.7\\\rowcolor{carolinablue!60}
\textbf{ALLaVA-Phi3} & \textbf{56.9} & 52.2 & \textbf{48.1} & 1382.3 & 32.7 & \textbf{667.8} & 53.0 & \textbf{347.1} & \textbf{32.9} & \textbf{37.8} & \textbf{41.1} & 64.0 & 68.5 & \textbf{54.8} & 68.1 & \textbf{55.3} & 69.0\\ 
\bottomrule
\end{tabular}

}
\label{tab:quantitative_results}
\end{table*}
\section{Conclusion}
In this work, we propose a Caption-then-QA, a simple-yet effective strategy to generate high-quality captions, instructions ans answers. We open-source the synthesized dataset ALLaVA, which is so far the largest dataset with 1.3M samples for LVLM training. Ablation studies show that our dataset is excelled at boosting performance for lite VLMs, making them on par with larger VLMs on multiple benchmarks.

\section{Limitation}
\label{sec:limitations}
Despite our significant efforts to enhance the quality of data, limitations still remain. 
Firstly, while \textit{Caption-then-QA} strategy exceeds directly generating responses in terms of accuracy (Section~\ref{sec:vflan_manual_inspection}), the answers are not completely accurate, indicating an area for ongoing refinement. 
Secondly, constrained by cost and effort, current version does not fully encompass multilingual content related to minority cultures. We anticipate that this issue can be addressed in future works, thereby enabling LVLMs to benefit a broader spectrum of groups.

\newpage





\bibliography{iclr2023_conference}
\bibliographystyle{iclr2023_conference}

\appendix
\clearpage

\section{Data Distillation}
\subsection{Deduplication on LAION Samples}
\label{app:dedup_laion}
For the Laion data, we used images from the Laion-400M~\citep{schuhmann2021laion400m} dataset. 
We filtered out high-resolution images with both width and height exceeding 512 pixels. 
Next, we excluded URLs potentially related to pornographic and violent content.
Finally, we applied a semantic retriever to deduplicate the images. Using all-mpnet-base-v2\footnote{https://huggingface.co/sentence-transformers/all-mpnet-base-v2} as the encoder, we obtained the semantic embeddings of the image captions. We then filtered out images whose caption embeddings dot product similarity smaller than 440 for deduplication.

\subsection{Prompt for Distilling LAION}
\label{app:prompt_laion}

\lstset{
 backgroundcolor=\color[RGB]{245,245,244},
 breaklines=true,
 basicstyle=\ttfamily\small
}
\begin{lstlisting}
### You are an excellent image describer and questioner
### You have three tasks in total
#### Your first task is to describe the given image as detailed as possible
#### Your second task is to ask a complex question that requires close inspection of the image and strong reasoning ability to answer, you should ask FIVE candidate questions in different aspects and diverse ways, then RANDOMLY choose one of them to answer
#### Your third task is to answer the question you raised solely based on the given image
### When you ask questions, try to find the most valuable information in the picture to ask about, and ask a question that is relevant to that information
### When you ask questions, do not involve violence, advertisement, possible invasion of privacy, or questions that may cause discomfort
### Do not mention anything from the prompt in your response
### You will follow the instructions to the best of your ability
### Your response should follow the following format
<start of description>
{description}
<end of description>
<start of candidate questions>
{candidate questions}
<end of candidate questions>
<start of question>
{question}
<end of question>
<start of answer>
{answer}
<end of answer>
\end{lstlisting}

\subsection{Prompt for Distilling Vision-FLAN}
\label{app:prompt_vflan}
\lstset{
 backgroundcolor=\color[RGB]{245,245,244},
 breaklines=true,
 basicstyle=\ttfamily\small
}
\begin{lstlisting}
You are an excellent image describer.

Your task is to first describe an image and then answer a question.

Your description should include details about the main subjects, background elements, colors, and any notable features. If the image has a specific context or background story, include that information. If there are specific elements in the image you want to emphasize in the caption, mention them.

Your answer should provide relevant information to the question and demonstrate the process of solving the question.

Both your description and answer should be professional, insightful, helpful, objective, unbiased. 

For scenarios where bias has been traditionally an issue, make sure that key traits such as gender and race are specified and in an unbiased way in the description -- for example, prompts that contain references to specific occupations.

If the question tries to induce you to produce something against ethical rules, such as leaking personal information or making discriminative judgements on underrepresented groups, you must point out the inappropriate intent and refuse to answer the question.

Here is the question:
```question
{question}
```

Your output should follow the format below:

<start of description>
{description}
<end of description>

<start of detailed answer>
{detailed_answer}
<end of detailed answer>
\end{lstlisting}


\subsection{Data Example}
\label{sec:data_example}

\begin{table*}[!t]
\scriptsize
\centering
\caption{Examples of our data. \textbf{Entries in bold} are generated by GPT-4V using our data generation protocol.}
\begin{tabular}{p{2.0cm}p{9cm}}
\toprule
\multicolumn{2}{l}{\textbf{LAION}}  \\
\midrule
& {\arraybackslash\includegraphics[width=.7in]{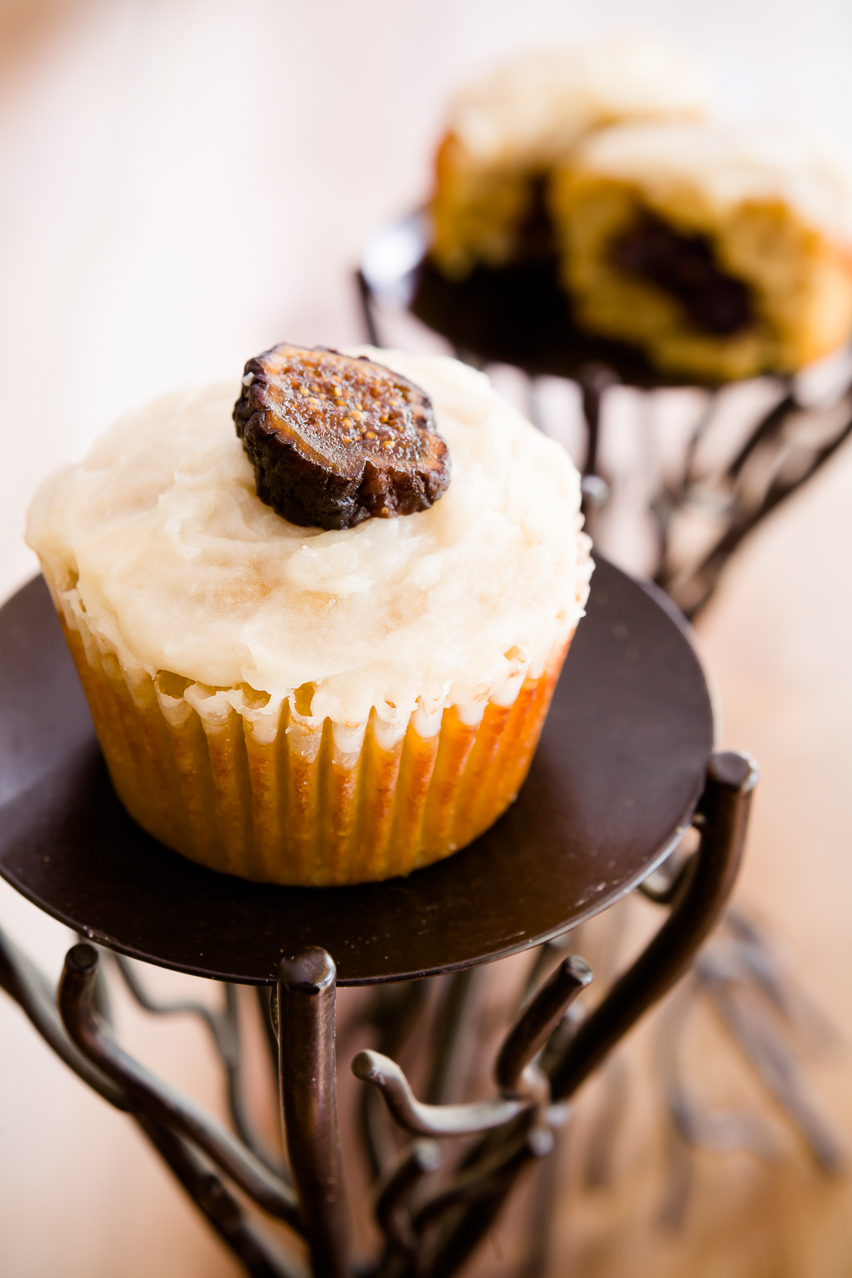}} \\
LAION Caption & Paleo Diet Fig Newton Cupcakes \\
\\
\textbf{Fine-grained Caption} & The image features a close-up view of a single cupcake with a creamy white frosting on top, placed on a dark-colored metal stand with slender, branch-like supports. The cupcake is centered in the frame and is positioned on a circular platform that is part of the stand. On top of the frosting, there is a garnish that appears to be a slice of candied fruit, possibly a fig, with visible seeds and a caramelized exterior, which adds a touch of elegance to the presentation. The cupcake liner has a fluted design and is a golden yellow color, suggesting it might be made of a buttery or vanilla-flavored cake. In the background, slightly out of focus, there is another cupcake with a bite taken out of it, revealing a dark filling, possibly chocolate or fruit jam. The background is a soft, neutral color, providing a contrast that highlights the cupcake in the foreground. \\
\\
\textbf{Complex Reasoning Instruction} & Considering the visual cues present in the image, what might be the flavor profile of the cupcake and how does the garnish on top contribute to the overall taste experience? \\
\\
\textbf{Detailed Answer} & Based on the image, the cupcake likely has a sweet and rich flavor profile, with the golden yellow cake suggesting a vanilla or buttery base. The creamy white frosting adds a layer of sweetness and a smooth texture. The garnish on top appears to be a slice of candied fig, which would contribute a chewy texture and a complex sweetness with hints of caramelization. The seeds within the fig slice would also add a slight crunch, creating a varied mouthfeel. Overall, the garnish not only serves as an aesthetically pleasing decoration but also enhances the flavor experience with its unique taste and texture. \\
\toprule
\multicolumn{2}{l}{\textbf{Vision-FLAN}}  \\
\midrule
& {\arraybackslash\includegraphics[width=1in]{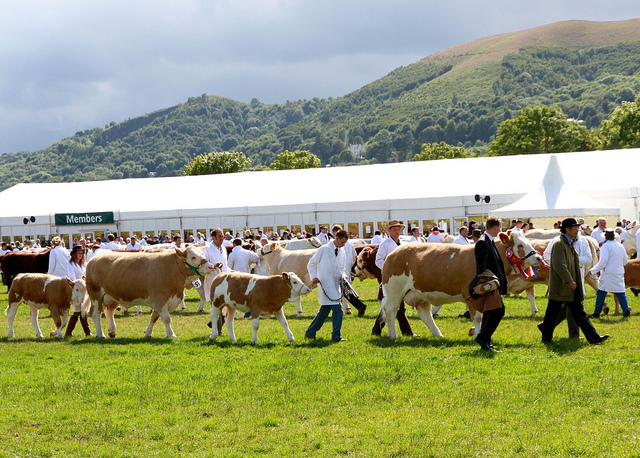}} \\
\textbf{Fine-grained Caption}& The image depicts a pastoral scene at what appears to be a livestock show. In the foreground, a row of individuals, primarily wearing white coats, are leading a line of large, tan and white animals across a grassy field. These animals, with their distinctive humped backs and white patches, are cattle, likely a specific breed given the context of the event. In the background, there's a large white tent labeled "Members", suggesting an exclusive area or services for participants or special guests. The landscape features rolling hills, and the sky is partly cloudy, allowing for patches of blue sky and sunlight to enhance the bucolic atmosphere. No other animal species are visible in the frame. \\
\\
Instruction & Given an image of an animal, identify the kind of animal in the image. The picture could be of more popular animals that are visible around zoos or are sometimes domesticated at home. They could also sometimes be found in the wild.
Options: (a) This image contains a giraffe (b) This image contains an elephant (c) This image contains a zebra (d) This image contains a sheep (e) This image contains a bear (f) This image contains a horse (g) This image contains a cow (h) This image contains a cat (i) This image contains a dog (j) This image contains a bird
 \\ 
\\
VFLAN Answer & (g) This image contains a cow \\
\\
\textbf{Detailed Answer} & The animals in the image are cattle, more commonly referred to as cows. This determination is made clear by their physical characteristics—large size, humped backs, distinctive color patterns, and the presence of horns in some individuals. Given the setting of a livestock show, these cows are likely to be a breed prized for either dairy or beef production. They are being shown off by handlers, which is typical in such events where animals are judged based on breed standards or other criteria. Therefore, the correct answer from the provided options is: (g) This image contains a cow. \\
\bottomrule
\end{tabular}
\label{tab:data_exp}
\end{table*}

In Table~\ref{tab:data_exp}, we show two examples from each of the datasets, with an image, a caption of the image, a question, and an answer. 
The \textbf{entries in bold} are generated by GPT-4V using our data generation protocol.

\subsection{Manual Inspection on Subsets of VFLAN}

\begin{figure}
    \centering
    \resizebox{\textwidth}{!}{
    \includegraphics{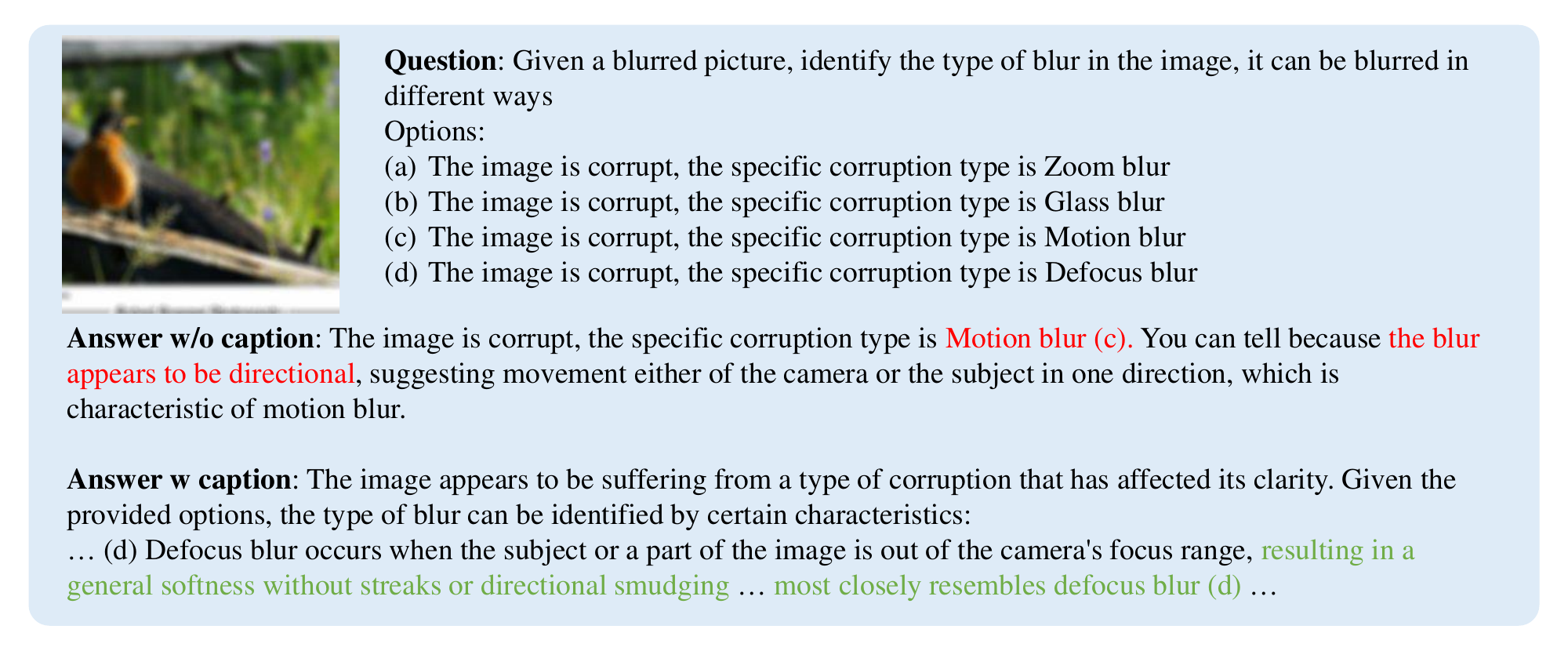}}
    \caption{Comparison between answer without captioning and answer with caption. The parts marked in \textcolor{red}{red} indicate errors in the answers, while the parts marked in \textcolor{green}{green} indicate the correct portions of the answers. These markings are used solely for better visualization purposes.}
    \label{fig:caption_advantage}
\end{figure}
\label{app:vflan_manual_inspection}
To verify the quality of the generated data, we uniformly sampled 100 pieces of data according to the original categories of vflan and conducted a detailed manual inspection. The inspection results are shown in Table~\ref{tab:vflan_manual_inspection}. During the inspection, we found that the method of captioning first and then answering, by providing a detailed description of the image beforehand, reduced the occurrence of hallucinations by the model. As shown in the Figure~\ref{fig:caption_advantage}, in the answers without prior captioning, the model incorrectly assumed that the blur in the image had directionality, thus concluding that the blur was caused by motion. However, in the case where captioning was done first and then answering, the model accurately identified the blur in the image as defocus blur, thanks to the help of the caption. This further demonstrates the advantage of captioning first and then answering.


\section{Details of Topic Analysis}

\subsection{Prompt for Topic Name Generation}
\label{app:prompt_gpt4_topic_summary}
Given a list of ten words per cluster, we leverage GPT-4 to generate a topic name based on these keywords with the prompt shown in Figure~\ref{fig:topic_prompt}.
\lstset{
 backgroundcolor=\color[RGB]{245,245,244},
 breaklines=true,
 basicstyle=\ttfamily\small
}
\begin{figure}[]
\begin{lstlisting}
I will provide you with a list of words obtained from LDA analysis. 
I want you to summarize the list of words using **1~2 words**.
You should only output the summary.
```list of words
{str(key_words)}
```
\end{lstlisting}
\caption{Prompt for generate topic names}
\label{fig:topic_prompt}
\end{figure}

\subsection{Full List of Topics}
\label{app:list_of_topics}
Table~\ref{tab:list_of_topics} shows a full list of topics generated by LDA and GPT-4.

\begin{table}
\centering
\caption{Full list of topics generated by LDA and GPT-4.}
\resizebox{\linewidth}{!}{
\begin{tabular}{|c|c|c|}
\hline
\textbf{Topic Name} & \textbf{List of Keywords} & \textbf{Proportion} \\
\hline
Scientific Research & cell, cost, class, phase, equation, sample, treatment, brain, decrease, variable & 4.1\%\\
\hline
Entertainment \& Collectibles & mask, costume, film, armor, figurine, smoke, movie, skateboard, moon, soldier & 4.0\%\\
\hline
Physical Appearance & fair, lip, finger, mustache, beard, complexion, nondescript, contemplative, indistinct, shoulder\_length & 4.0\%\\
\hline
Artistic Writing & ribbon, bow, script, decorate, cursive, lettering, handwritten, multicolore, stamp, bubble & 3.9\%\\
\hline
Fashion Accessories & diamond, sheen, pendant, zipper, bead, backpack, earring, weave, compartment, pad & 4.0\%\\
\hline
Geometric Orientation & triangle, centrally, marker, occupy, outer, orient, diagonal, triangular, providing, website\_address & 3.9\%\\
\hline
Hardware Components & hole, blade, mechanical, screw, grip, temperature, connector, port, assembly, slot & 4.0\%\\
\hline
Nature Seasons & snow, forest, sunlight, rugged, canopy, sparse, autumn, bare, spring, desert & 4.0\%\\
\hline
Urban Environment & truck, bus, sidewalk, overhead, pole, pedestrian, shop, bustle, signage, coin & 4.0\%\\
\hline
Artistic Drawing & sketch, monochromatic, elephant, train, drawing, feather, minimalistic, artistic, realistic, grayscale & 4.0\%\\
\hline
Vehicle Parts & motorcycle, bike, bicycle, tire, hood, rim, headlight, rubber, rider, damage & 3.9\%\\
\hline
Food Ingredients & slice, dish, cake, ingredient, chocolate, vegetable, cheese, apple, pizza, flavor & 4.2\%\\
\hline
Sports \& Aviation & fuselage, military, stadium, baseball, registration, sock, football, airplane, jet, identification & 3.9\%\\
\hline
Professional Achievements & headline, publication, entity, award, certificate, political, research, skill, paragraph, strategy & 4.1\%\\
\hline
Kitchen Items & countertop, cabinet, sink, jar, lid, knob, keyboard, liquid, rack, basket & 3.9\%\\
\hline
Home Decor & carpet, lamp, vase, string, sofa, curtain, sheet, guitar, cozy, instrument & 4.0\%\\
\hline
Nature \& Biology & petal, elongate, stem, organic, tiny, butterfly, translucent, fluffy, delicate, breed & 4.0\%\\
\hline
Technology Services & lens, tablet, mobile, customer, app, audio, template, ticket, channel, log & 4.1\%\\
\hline
Fashion Items & sleeveless, wrist, skirt, trouser, garment, knee, jean, sneaker, heel, neckline & 4.1\%\\
\hline
Artistic Items & desk, vest, tattoo, discern, nondescript, indistinct, sticker, artist, notebook, difficult & 3.9\%\\
\hline
Historical Artifacts & classical, sculpture, statue, crown, th\_century, shield, dial, robe, religious, inscription & 4.0\%\\
\hline
Sports Event & horse, spectator, tank, crowd, basketball, athletic, athlete, court, midst, intense & 4.0\%\\
\hline
Gardening Construction & garden, bench, rail, palm, flank, greenery, nail, tropical, construct, plank & 4.0\%\\
\hline
Event Celebration & boy, girl, outdoors, pool, venue, celebration, bouquet, joyful, wedding, member & 4.0\%\\
\hline
Natural Scenery & tower, river, early\_morne, sand, sunlight, dusk, wind, ocean, tranquil, late\_afternoon & 4.0\%\\
\hline
\end{tabular}
}
\label{tab:list_of_topics}
\end{table}

\section{Training details}
\label{app:training_details}

\subsection{Training data}
\label{app:training_data}
Table~\ref{tab:training_dataset_by_stage} summarizes the training data for ALLaVA models.

\begin{table*}[h]
\centering
\small
\caption{Training datasets for ALLaVA models. 
We include OpenChat~\citep{wang2023openchat} in \textit{Text}, ShareGPT4V~\citep{chen2023sharegpt4v} in \textit{Caption} and {llava\_instruct\_657K}~\citep{liu2023improved} in \textit{VQA}.
}
\begin{tabular}{l | l | lrr | r |r}
\toprule
Stage & Type & Name & \#Ex. & \#Epoch & Type Total & Stage Total \\
\midrule
 \multirow{4}{*}{PT.} & \multirow{2}{*}{\textit{Text}} & \textbf{\textit{Evol-Instruct-GPT4-Turbo-143K}} & 143K & \multirow{2}{*}{2} & \multirow{2}{*}{298K} & \multirow{4}{*}{1,042K} \\
 & & OpenChat & 6K & &  \\ \cline{2-6} 
 & \multirow{2}{*}{\textit{Caption}} & \textit{\textbf{ALLaVA-Caption-4V}} & 664K & \multirow{2}{*}{1} & \multirow{2}{*}{744K} \\
 & & ShareGPT4V & 80K & & \\ \cline{1-7}
  \multirow{4}{*}{FT.} & \multirow{2}{*}{\textit{Text}} & \textbf{\textit{Evol-Instruct-GPT4-Turbo-143K}} & 143K & \multirow{2}{*}{1} & \multirow{2}{*}{149K} & \multirow{4}{*}{1,469K} \\
 & & OpenChat & 6K & &  \\ \cline{2-6} 
 & \multirow{2}{*}{\textit{VQA}} & \textit{\textbf{ALLaVA-Instruct-4V}} & 663K & \multirow{2}{*}{1} & \multirow{2}{*}{1,320K} \\
 & & llava\_instruct\_657K & 657K & & \\ \bottomrule
\end{tabular}
    \label{tab:training_dataset_by_stage}
\end{table*}

\subsection{Hyperparameters}
We detail the training hyperparameters for ALLaVA models in Table~\ref{tab:hyperparams}.

\begin{table}[t]
\centering
\caption{Training hyperparameters.}
\begin{tabular}{c | c c}
\toprule
Stage & Name & Value \\
\midrule
 \multirow{9}{*}{1} & Global Batch Size & 256 \\ 
 & Deepspeed ZeRO Stage & 1  \\
 & Optimizer & AdamW  \\
 & Weight Decay & 0  \\
 & Scheduler & Cosine Annealing with Linear Warmup  \\
 & Warmup Ratio & 0.03  \\
 & Max LR & $2e-5$  \\
 & Min LR & $2e-6$  \\
 & Precision  & BF16  \\ \midrule
\multirow{9}{*}{2} & Global Batch Size & 128 \\ 
 & Deepspeed ZeRO Stage & 1  \\
 & Optimizer & AdamW  \\
 & Weight Decay & 0  \\
 & Scheduler & Cosine Annealing with Linear Warmup  \\
 & Warmup Ratio & 0.03  \\
 & Max LR & $2e-5$  \\
 & Min LR & $2e-6$  \\
 & Precision  & BF16  \\
 \bottomrule
\end{tabular}
\label{tab:hyperparams}
\end{table}

\subsection{Computation Resources}
\label{app:computation_resources}
All experiments are conducted on a single 8*A800 GPU node. We show the training time of each model in Table~\ref{tab:computation_resources}.
\begin{table}[h]
    \centering
    \caption{Training time (in hour) for each ALLaVA model on a single 8*A800 GPU node. }
    \begin{tabular}{l|cc}
    \toprule
         Model& PT. & FT. \\
    \midrule
         ALLaVA-Phi2 & 8.3 & 10.6 \\
         ALLaVA-StableLM2 & 4.0 & 11.2 \\ 
         ALLaVA-Phi3 & \textasciitilde 18 &  \textasciitilde 24 \\ 
         \bottomrule
    \end{tabular}
    \label{tab:computation_resources}
\end{table}

\section{Evaluation}

\subsection{Details on Benchmarks}
\label{sec:benchmarks}




The benchmarks employed in this study are detailed below.

Vicuna-80~\citep{vicuna2023}; 
GQA~\citep{hudson2019gqa};
HallB: HallusionBench~\citep{guan2023hallusionbench};
MME (Perception and Cognition)~\citep{fu2023mme};
MMVP~\citep{tong2024eyes};
TS: TouchStone~\citep{bai2023touchstone}; 
VQA$^T$: TextVQA~\citep{singh2019towards};
MV: MathVista~\citep{lu2024mathvista};
MM-Vet~\citep{yu2023mmvet};
MMMU$^{val}$: validation set of MMMU~\citep{yue2023mmmu};
SQA$^{I}$: ScienceQA-Image~\citep{lu2022learn}; 
LLaVA$^{W}$: LLaVA-Bench (In-the-Wild)~\citep{liu2023visual}; 
MB: MLLM-Bench~\citep{ge2023mllmbench};
MMB: MMBench~\citep{liu2023mmbench};
SEED\(_{img}^{v1}\): image set of SEED-Bench-v1~\citep{li2023seedbench}; 

\begin{itemize}
\item GQA~\citep{hudson2019gqa} consists of 12,578 questions for real-world reasoning and compositional question answering. \textit{Accuracy} is used as the metric.

\item HallB: HallusionBench~\citep{guan2023hallusionbench} is composed of 254 samples for evaluating the hallusination problem of LVLMs. \textit{Accuracy} is used as the metric.

\item MME~\citep{fu2023mme} is a benchmark with 2,374 questions spanning 14 subtasks. \textit{Accuracy} is used as the metric.

\item MMVP~\citep{tong2024eyes} is a benchmark aiming to test ``CLIP-blind'' pairs.  \textit{Accuracy} is used as the metric.

\item TouchStone~\citep{bai2023touchstone} contains 908 open-ended question covering 5 abilities and 27 subtasks. LVLM's answers are compared with pre-generated text-based GPT-4's answers, using text-based GPT-4 as the judge to score each answer. The \textit{Averaged Scores} are used as the metric.

\item TextVQA~\citep{singh2019towards} comprises 5,000 questions and \textit{Accuracy} is used as the metric.

\item MathVista~\citep{lu2024mathvista} consists of 6,141 samples aiming to test the mathmatical reasoning ability of LVLMs.

\item MM-Vet~\citep{yu2023mmvet} comprises 218 questions, each requiring multiple capabilities to solve and provided with multiple groundtruths for a flexible match. \textit{Accuracy} is adopted as the metric.

\item MMMU~\citep{yue2023mmmu} (val set) consists of 900 multiple-choice questions that require expert-level knowledge to solve. \textit{Accuracy} is adopted as the metric.

\item ScienceQA~\citep{lu2022learn} contains 4,201 questions encompassing different subjects and categories. \textit{Accuracy} is adopted as the metric.

\item LLaVA-Bench (In-the-Wild)~\citep{liu2023visual} contains 60 open-ended questions and uses text-based GPT-4~\citep{openai2023gpt4} as a judge to score answers in a pairwise fashion. \textit{Score Ratio} between candidate answers and anchor answers from GPT-4 is adopted as the metric.

\item MLLM-Bench~\citep{ge2023mllmbench} contains 420 complex visual reasoning questions along with per-sample criteria that aids GPT-4V to score each answer. LLaVA-v1.5-13B's answers serve as anchors. \textit{Win rate over the anchor} is adopted as the metric.

\item MMBench~\citep{liu2023mmbench} (dev set) consists of 4,329 multiple-choice questions across 20 ability dimensions, using \textit{Accuracy} under circular evaluation as the metric.

\item SEED-Bench-v1~\citep{li2023seedbench} (image set) comprises 14,233 multiple-choice questions across 9 dimensions. \textit{Accuracy} is adopted as the metric.


\end{itemize}

\subsection{Evaluation Prompt for Vicuna-80}\label{app:vicuna_prompt}
\lstset{
 backgroundcolor=\color[RGB]{245,245,244},
 breaklines=true,
 basicstyle=\ttfamily\small
}
\begin{lstlisting}
You are a fair judge. You will be provided with a question and two answers. Your task is to judge which answer is of better quality.
Here is the question:
{question}

Here is Answer1:
{answer1}

Here is Answer2:
{answer2}

Your output should be either "Answer1" or "Answer2".
\end{lstlisting}

\section{Acknowledgement}
\label{sec:acknowledgement}
ALLaVA dataset is for research purpose only. Please carefully check the licenses of the original datasets (\href{https://vision-flan.github.io/}{VFLAN} and \href{https://laion.ai/blog/laion-400-open-dataset/}{LAION-400M}) before using ALLaVA as we do not own the images. The images may be taken down at any time when requested by the original dataset owners or owners of the referenced images. 
 If you hope to take down any images, please raise an issue on our \href{https://huggingface.co/datasets/FreedomIntelligence/ALLaVA-4V}{Huggingface Repository}.

\end{document}